\documentclass[sigconf]{acmart}

\usepackage{url}
\usepackage{amsmath,amsfonts}
\usepackage{accents}
\usepackage[linesnumbered,ruled]{algorithm2e}
\usepackage[noend]{algpseudocode}
\usepackage{graphicx}
\usepackage{textcomp}
\usepackage{comment}
\usepackage{multicol,multirow}
\usepackage[T1]{fontenc}
\def\BibTeX{{\rm B\kern-.05em{\sc i\kern-.025em b}\kern-.08em
    T\kern-.1667em\lower.7ex\hbox{E}\kern-.125emX}}
\usepackage{todonotes}
\usepackage{caption}
\usepackage{wrapfig}
\usepackage{subfig}
\newsavebox{\measurebox}
\usepackage{threeparttable}
\usepackage[font=small]{caption} 
\usepackage{enumitem}
\usepackage{booktabs}

\usepackage{hyperref}
\hypersetup{
    colorlinks=false
}

\makeatletter
\newcommand{\algorithmfootnote}[2][\footnotesize]{%
  \let\old@algocf@finish\@algocf@finish
  \def\@algocf@finish{\old@algocf@finish
    \leavevmode\rlap{\begin{minipage}{\linewidth}
    #1#2
    \end{minipage}}%
  }%
}
\makeatother

\newcommand*{\Scale}[2][4]{\scalebox{#1}{$#2$}}%

\AtBeginDocument{%
  \providecommand\BibTeX{{%
    Bib\TeX}}}

\setcopyright{acmcopyright}
\copyrightyear{2023}
\acmYear{2023}
\acmDOI{XX.XXXX/XXXXXXX.XXXXXXX}

\acmConference[CHASE '23]{EEE/ACM international conference on Connected Health: Applications, Systems and Engineering Technologies}{June 21-23,
  2023}{Orlando, FL}
\acmPrice{15.00}
\acmISBN{978-1-4503-XXXX-X/18/06}

\begin{document}

\title[short title]{Short: Basal-Adjust: Trend Prediction Alerts and Adjusted Basal Rates for Hyperglycemia Prevention}

\author{Chloe Smith}
\email{cas8ds@virginia.edu}
\affiliation{%
  \institution{University of Virginia}
  \city{Charlottesville}
  \state{Virginia}
  \country{USA}
  \postcode{22904}
  }

\author{Maxfield Kouzel}
\email{mak3zaa@virginia.edu}
\affiliation{%
  \institution{University of Virginia}
  \city{Charlottesville}
  \state{Virginia}
  \country{USA}
  \postcode{22904}
}

\author{Xugui Zhou}
\email{xugui@virginia.edu}
\affiliation{%
  \institution{University of Virginia}
  \city{Charlottesville}
  \state{Virginia}
  \country{USA}
  \postcode{22904}
}

\author{Homa Alemzadeh}
\email{ha4d@virginia.edu}
\affiliation{%
  \institution{University of Virginia}
  \city{Charlottesville}
  \state{Virginia}
  \country{USA}
  \postcode{22904}
}

\begin{abstract} 
Significant advancements in type 1 diabetes treatment have been made in the development of state-of-the-art Artificial Pancreas Systems (APS). However, lapses currently exist in the timely treatment of unsafe blood glucose (BG) levels, especially in the case of rebound hyperglycemia. We propose a machine learning (ML) method for predictive BG scenario categorization that outputs messages alerting the patient to upcoming BG trends to allow for earlier, educated treatment. 
In addition to standard notifications of predicted hypoglycemia and hyperglycemia, we introduce BG scenario-specific alert messages and the preliminary steps toward precise basal suggestions for the prevention of rebound hyperglycemia.
Experimental evaluation on the DCLP3 clinical dataset achieves >98\% accuracy and >79\% precision for predicting rebound high events for patient alerts.
\end{abstract}

\ccsdesc[500]{Computer systems organization~Dependable and fault-tolerant system; Embedded and cyber-physical systems}
\ccsdesc{Applied computing~Life and medical sciences}
\ccsdesc[100]{Computing Methodologies~Machine learning; Modeling and simulation}

\keywords{
Safety, Cyber-Physical Systems, Medical Device, Wearable Systems, Body Sensor Networks.}

\maketitle
\thispagestyle{plain}
\pagestyle{plain}

\section{Introduction}



Patients with type 1 diabetes monitor and treat their BG levels, as their pancreases do not produce the required levels of insulin. 
Recent diabetes management research largely focuses on the development of APS that combine insulin pump and CGM technology \cite{closedSupport}. One current task is the improvement of APS controllers \cite{study1,study2,study3} or design of monitors \cite{zhou2021data,zhou2023hybrid,zhou2022robustness} to detect and prevent hyperglycemia \cite{hyper} and hypoglycemia \cite{mayolow}, which in mild cases can cause blurred vision, weakness or confusion, and in extreme cases result in coma or even death. 
One of the limitations of the existing APS is in addressing rebound highs/lows \cite{reboundCGM}.
These are typical scenarios for type 1 diabetes patients where after treating an episode of out-of-range BG levels, BG shifts through the target range to the opposite unsafe range (i.e., low to high or vice versa). 
Current APS can alert users about upcoming trends and rebound highs/lows \cite{fdaGuardian}, but they neglect to take or advise precise treatment actions.

In this paper, we propose methods for the prediction and prevention of rebound highs which can be integrated with the existing APS to address the following two concerns.
First, patients may take action too late to prevent their BG levels from going outside the safe range. 
To address this case, we propose a predictive alert mechanism that uses an ML algorithm to estimate BG levels an hour prior, categorizes BG trends into one of the several BG scenarios (e.g., \textit{rebound high}), and alerts the user of both general high/low BG levels and specific rebound scenarios.
Since carbohydrate (carb, CHO) ingestion and insulin take time to affect BG levels, earlier action can mitigate unsafe BG levels and help users stay within the safe range. 
Second, low BG levels may be over-corrected. We introduce a method to recommend increases in basal injections given via insulin pump when BG has shifted to the safe range, based on carb input entered when BG was low. This adjusted basal output accounts for earlier
carb inputs once in the safe range, preventing a rebound high from occurring.
Addressing these two concerns can help to attain in-range BG values. 

Our contributions are as follows: 
\vspace{-0.5cm}
\begin{enumerate}
    \item Predictive alert mechanism based on an ML prediction model for BG scenario-specific categorization and messaging
    \item Preliminary steps towards developing a method for delayed increased basal recommendations to patients for preventing hyperglycemia based on prior carb intake
    \item Evaluation using a realistic testbed's simulated data and actual patient data
\end{enumerate}


\section{Background}
\textbf{Artificial Pancreas Systems:}
\begin{figure}
        \centering
    \includegraphics[scale=.38]{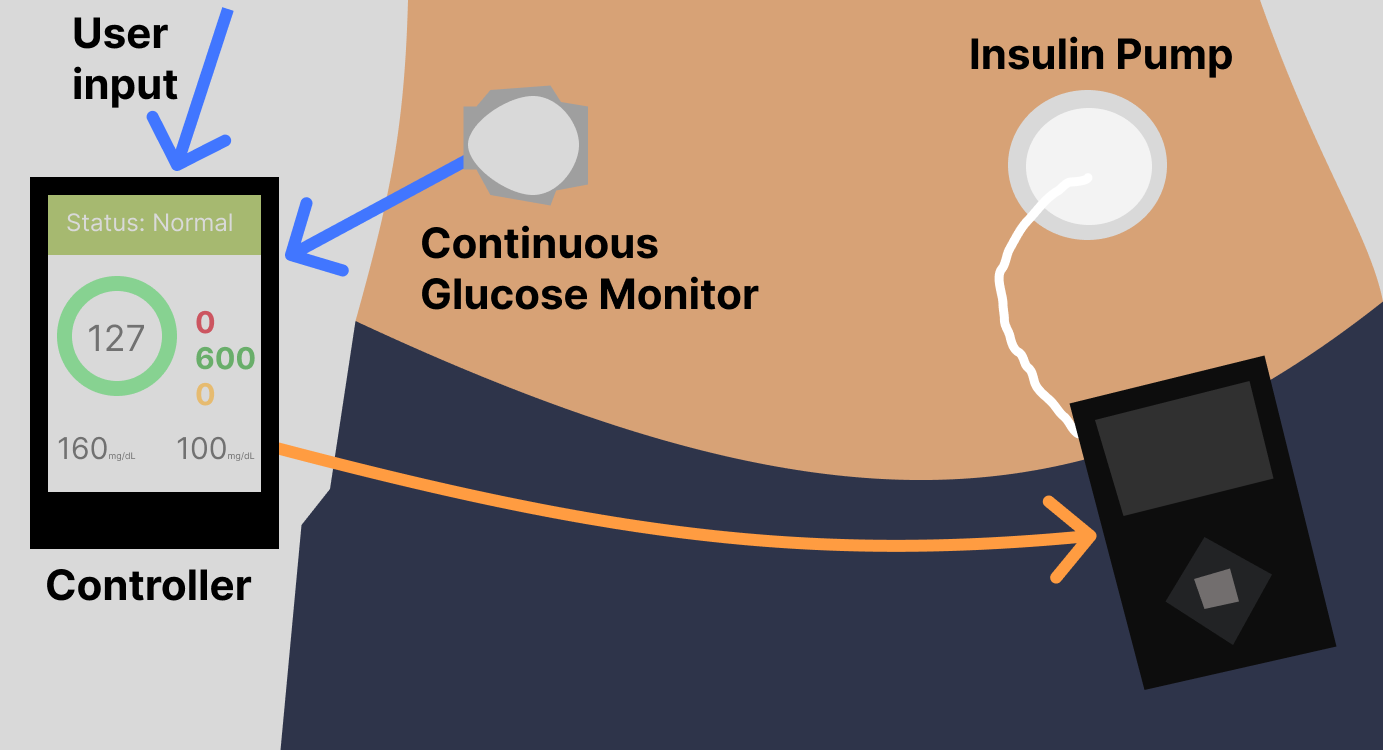} \\
        \vspace{-1em}
        \caption{Artificial Pancreas System}
        \label{fig:APS}
        \vspace{-1.5em}
    \end{figure}
As shown in Figure \ref{fig:APS}, APS are closed-loop systems that control BG levels using three devices: (i) a CGM that detects BG levels, (ii) a controller that intakes these BG levels (and user input, such as carbs) and determines insulin pump instructions (dose), and (iii) an insulin pump which injects instructed insulin doses. 
We focus on APS that allow for standard carb input. BG levels are also impacted by temporary basal rate adjustments (e.g., for exercise), extended boluses (e.g., for fat-rich meals), or consuming carbs (e.g., for low BG).

\textbf{Basal: }
Basal is an insulin injection often taken at a small, hourly rate to mimic a pancreas' gradual release to prevent high BG.

\textbf{BG Safe Ranges: }
The safe range for BG levels, in which no negative symptoms exist, is defined to between 70 and 180 $mg/dL$. 
\emph{Hypoglycemia} refers to BG levels below 70 $mg/dL$, and \emph{hyperglycemia} refers to BG levels above 180 $mg/dL$.

\textbf{Over-correction of Low BG: }
Over-correction of low BG occurs when consumption of carbs is beyond what is required to draw BG levels up from hypoglycemia to the safe range. 
In cases of hypoglycemia, treatment raises BG levels by consuming a set number of fast-acting carbs in set intervals until BG is back in range \cite{mayolow}. This is less precise than a correction insulin dose for hyperglycemia, which is calculated rather than set.
In addition, meals often possess more carbs than are needed to treat hypoglycemia, and insulin dosing to account for excess carbs when in this state can be risky.

\textbf{Rebound High/Low: }
Rebound hyperglycemia, or "a rebound high," is defined in previous works as a "series of sensor glucose values (SGVs) >180 $mg/dL$ starting within two hours of an antecedent SGV <70 $mg/dL$" \cite{reboundCGM}. We define a rebound low as the reverse, a series of sensor glucose values (SGVs) <70 $mg/dL$ starting within two hours of an antecedent SGV <180 $mg/dL$. 
"Rebound high" can also refer to the Somogyi effect \cite{Somogyi}, 
an increase in blood sugar in the morning as a "rebound" from low blood sugar overnight, typically caused by natural hormone releases \cite{defDawnSomogyi}.
Although the Somogyi effect \emph{is} a rebound high, 
in this paper, we refer to "rebound highs" that occur due to over-correction of low BG, and not the rebounds that occur without intervention. 

\section{Related Work}

\textbf{Predictive BG Alerts:} In recent years, popular type 1 diabetes management device manufacturers have rolled out predictive BG messaging and alerting systems. Medtronic's Guardian Connect System 
claims to predict BG 10-60 minutes in advance and inform the patient of preemptive treatments \cite{fdaGuardian}. 
New data-driven technologies 
can also predict BG accurately for a maximum of 2 hours in advance \cite{Taisa2020,Dutta2018}. However, these works do not consider more complex scenarios, such as multiple varying meal inputs outside of night hours. 
They usually only alert on hyperglycemia and hypoglycemia events and also lack precise treatment advice and labeled scenarios (e.g., "rebound high").

\textbf{Rebound Hyperglycemia:}
Much of our work focuses on rebound hyperglycemia. A recent study evaluates the impact of existing nonspecific predictive alerts to notify patients of hypoglycemia on rebound high frequency \cite{reboundCGM}.
However, similar limitations exist in this rebound-focused work, as this study did not use alerts that specifically recognize rebound highs, and the alerts did not provide direct recommendations for insulin dosage or other user action.

    To address such limitations, we propose a BG prediction model that provides specific 
    scenario alerts and gives precise insulin recommendations for a rebound high.
    
\section{Methods}
\begin{figure}[t!]
    \centering
\includegraphics[scale=.45]{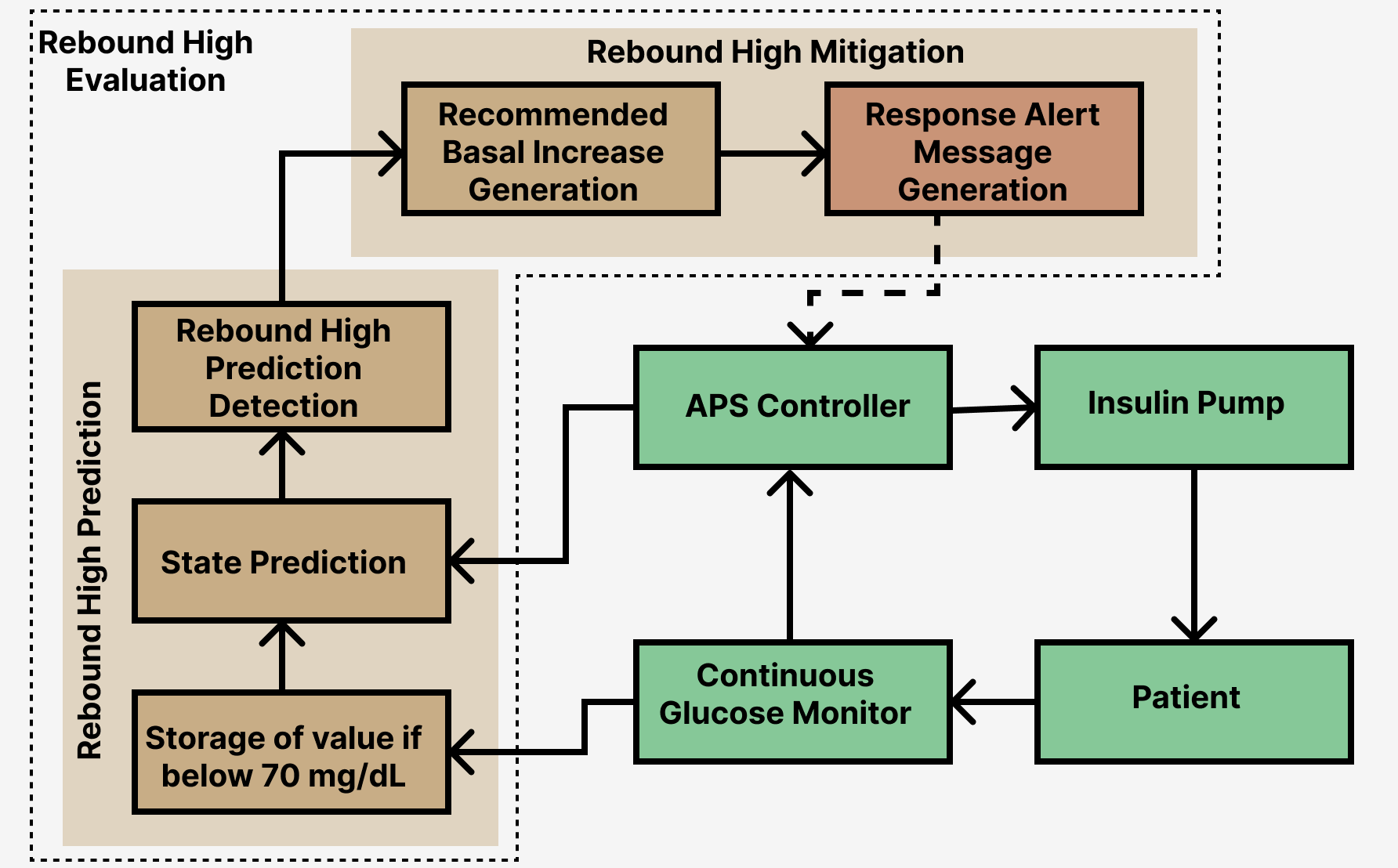} \\
    \vspace{-1em}
    \caption{Overall Design of Rebound High Evaluation Module}
    \label{fig:reboundHighArch}
    \vspace{-1.5em}
\end{figure}
We propose to supplement the existing APS controllers with mechanisms for preventing rebound hyperglycemia events. 
As shown in Figure \ref{fig:reboundHighArch}, our proposed Rebound High Evaluation module is designed using
ML-based APS state prediction, a recommended increased basal insulin to address carb input, and response alert message generation to assist patients in the detection and mitigation of rebound highs. 

\subsection{ML Model for Blood Sugar Prediction}\label{workflowdata}
\begin{figure}[b!]
    \vspace{-1em}
    \centering
\includegraphics[scale=.2]{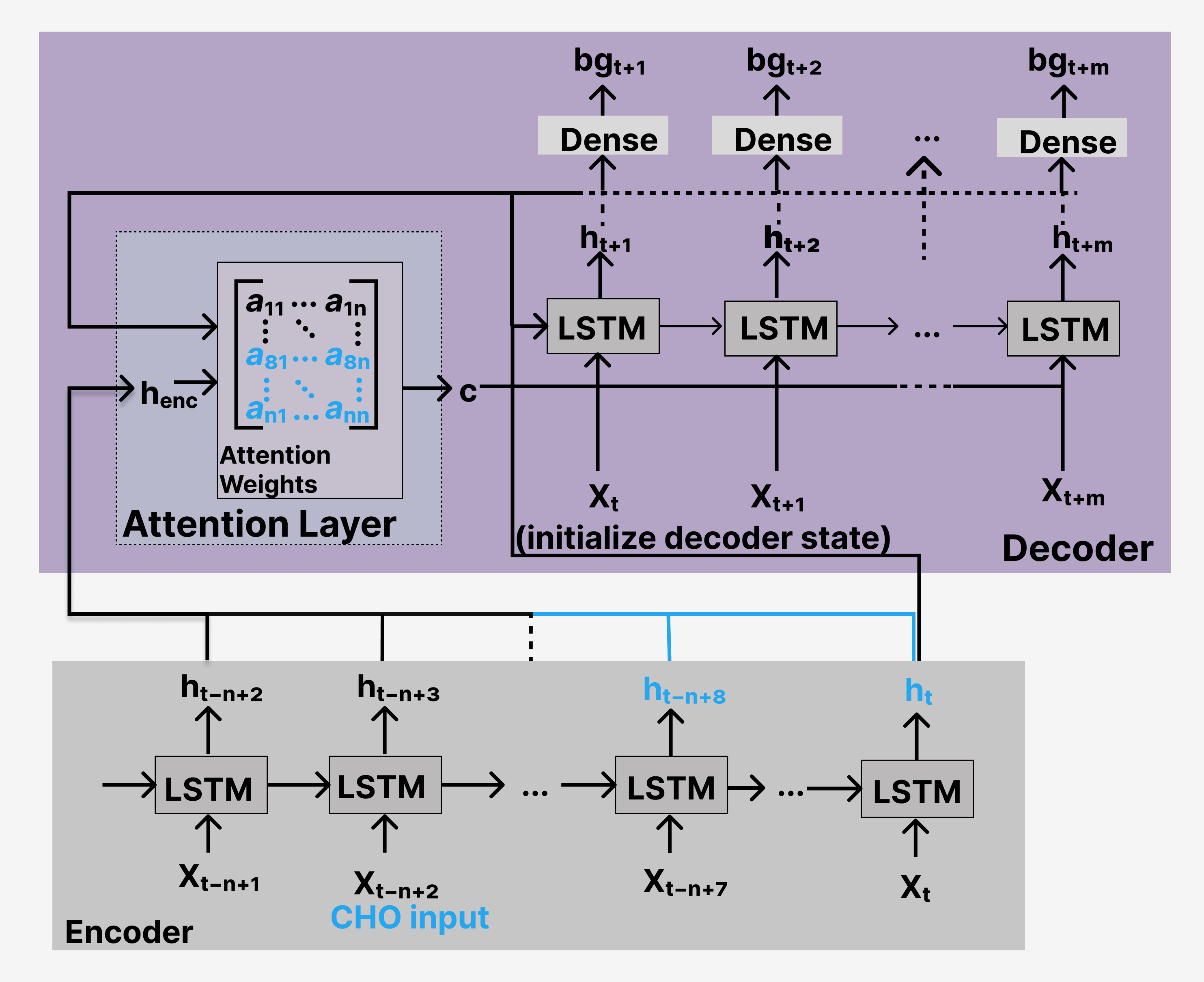} \\
    \vspace{-1em}
    \caption{ML Model for Blood Sugar Prediction}
    \label{fig:encodeDecode}
\end{figure}

To forecast future BG levels and detect trends, we design an encoder-decoder ML model with attention that outputs the expected BG values given the APS states as input (see Figure \ref{fig:encodeDecode}). Each state contains BG, insulin dosage, insulin on board (IOB), and carbs. The encoder generates a deep representation of the sequence of input states that the decoder uses to initialize its internal state as it predicts the BG value for the next timestep. 

We chose this architecture due to its success in sequence generation tasks \cite{hochreiter1997, BahdanauAttn}.
Unlike other BG prediction models that give a single estimate of BG at a fixed prediction horizon (e.g. 30 or 60 minutes), the encoder-decoder model generates a full sequence of BG predictions up to and including the prediction horizon where each previous prediction is factored into the next prediction autoregressively. This allows for the detection of the earliest possible occurrence of an adverse event, such as a rebound high. 

The attention layer has also been shown to help the decoder process the information learned by the encoder during prediction by amplifying relevant features, suppressing unimportant ones, and improving performance of encoder-decoder models in a variety of application scenarios \cite{BahdanauAttn, Zhu2022, Muralidhar2019}. Finally, it followed to use a recurrent neural network as the underlying layer in the encoder and decoder because the APS data has temporal order, so we chose the LSTM to retain past information over longer input sequences.

The following equations describe the overall model:
\begin{align}
    h_{enc} &= Encoder(X_{[t-n+1, t]}) \\
    s_{t'+1} &= (h_{t'+1}, z_{t'+1}) = LSTM(x_{t'}, s_{t'}, c_{t'+1}) \\
    c_{t'+1} &= A \cdot h_{enc} \\
    A &= \text{softmax}\left( w_v \cdot \text{tanh} ( W_q h_{t'} + W_k h_{enc} ) \right) \\
    \widehat{bg}_{t'+1} &= f(h_{t'+1})
\end{align}
\noindent where $X_{[t-n+1, t]}$ is the past $n$ APS states, and $h_{enc}$ represents the input learned by the encoder for all timesteps. In the decoder, described by equations (2)-(5), $x_{t'}$ is the projected APS state for future time $t'$, $s_{t'}$ is the decoder's internal state made of a hidden state $h_{t'}$ and a cell state $z_{t'}$, and $c_{t'}$ is the context vector generated by the attention layer using the attention weight matrix $A$. The decoder's hidden state $h_{t'}$ is passed through a feedforward neural network, $f$, to generate the predicted BG value, $\widehat{bg}$. The decoder calculates $s_{t'}$ for $t' \in [t+1, t+m]$, where $m$ is the length of BG predictions. $W_q$, $W_k$, and $w_v$ are weights tuned in the attention layer.

The attention layer within the decoder calculates how much the decoder should weigh each part of the input data (i.e., how much attention it should give) and uses a feedforward neural network to generate attention weights \cite{BahdanauAttn}. To help the model learn the delayed action of carbs on BG, which need 15-30 minutes to start impacting BG \cite{30min, freeman2009}, we customize the attention layer to increase the attention weight for input states 30 minutes after a carb input by 10\%. Forcing the model to focus on these states helps it learn  to account for the physiological delay that the meal carbs take to affect BG. 
\subsection{Recommended Increased Basal for Carb Carry-over}
We propose a method to find a basal rate increase for when carbs are entered into the APS by patient while 
hypoglycemic. We calculate an increased basal (IB) using the existing carb input data in the Basal-Bolus controller (typically used for meal bolus generation). We then recommend the IB to be applied once the BG level has returned to a safe range.

Specifically, our goal is to find a new, corrected basal rate $IB$ that is $\delta$ units of insulin greater than the current basal rate, $r_t$, at time $t$, 
such that all predicted BG values within the hour $BG_{[t,t+12]}$ are in range:
\begin{equation}
    IB = r_t + \delta \Rightarrow 70 < BG_{[t,t+12]} < 180
\end{equation}

 In this paper, we use domain knowledge and heuristics to choose a set of fixed $\delta$ values and select the one that minimizes timesteps outside of the BG safe range in predictions without inducing hypoglycmia.  We then add this $\delta$ to the normal basal rate to find an $IB$ to be suggested in the alerts. More accurate optimization of IB is the subject of future work and beyond the scope of this paper.
\subsection{Rebound High Evaluation and Alert Generation}
\begin{table*}[]
    \centering
    \caption{Rebound High Alert System Performance Metrics}
    \vspace{-0.5em}
    \vspace{-0.5em}
    \resizebox{\textwidth}{!}{%

    \begin{tabular}{cll} \toprule
        &\textbf{BG Scenario} & \textbf{Alert Message} \\
        \midrule
        1 & $((x_{p} > 180)\wedge(t=p-24,...,p|x_t \not < 70)$ & "Your blood sugar is predicted to increase to 180 mg/dL in the next hour.” \\
        2 & $((x_{p} > 180)\wedge(t=p-24,...,p|x_t < 70)$ & “You are predicted to have a rebound high in the next hour.  \\
        &&  Give a temp basal of $IB$ once you are back in range.”  \\
        3 & $((x_{p} < 70)\wedge(t=p-24,...,p|x_t \not > 180)$ & “Your blood sugar is predicted to decrease to 70 $mg/dL$ in the next hour.”  \\
        4 & $((x_{p} < 70)\wedge(t= p-24,...,p|x_t > 180)$ & "You are predicted to have a rebound low in the next hour. Suspend insulin temporarily." \\
        5 & else & - no message - \\
        \bottomrule
        \vspace{-1em}
    \end{tabular}
    }
    $x_t$ represents predicted BG level at a given timestep $t$, $p$ is the most up-to-date predicted timestep, $IB$ is the calculated Increased Basal value.
    \label{tab:alert_messages}
    \vspace{-1em}
\end{table*}
The prediction model's forecasted BG sequence can be used to detect hyperglycemia, hypoglycemia, and rebound events in advance. Predicted and past BG states are categorized to output appropriate alert messages as shown in Table \ref{tab:alert_messages}.
If BG is predicted to be out of range without being a rebound (rows 1, 3), then general alerts are issued.
If predicted BG is high within two hours of being low (row 2), then a rebound high alert with recommended IB is issued.
If predicted BG is low within two hours of being high (row 4), then a rebound low alert with recommended basal suspension is issued.
These messages include scenario-specific language for the rebound cases and provide specific suggestions.

We specifically define the following contextual conditional event to trigger an alert for a rebound high:

\vspace{-1em}
{
}%

\begin{equation}    
\Scale[0.99]{y_t = P(\exists t-12<l\leq t < h\leq t+12: BG_{l}<70 \wedge BG_{h}>180)}
%
\end{equation}
\noindent where $t$ is the current time and $l$, $h$ are APS timesteps. An alert triggers when $y_t \ge 0.5$. $BG$ is the sequence of BG values over 2 hours: (1) observed values within the previous hour and (2) predicted values for the next hour from the prediction model. This allows for the detection of rebound highs up to 1 hour in advance.

\section{Experimental Evaluation}
\subsection{Datasets}
\begin{figure}
        \centering
    \includegraphics[scale=.45]{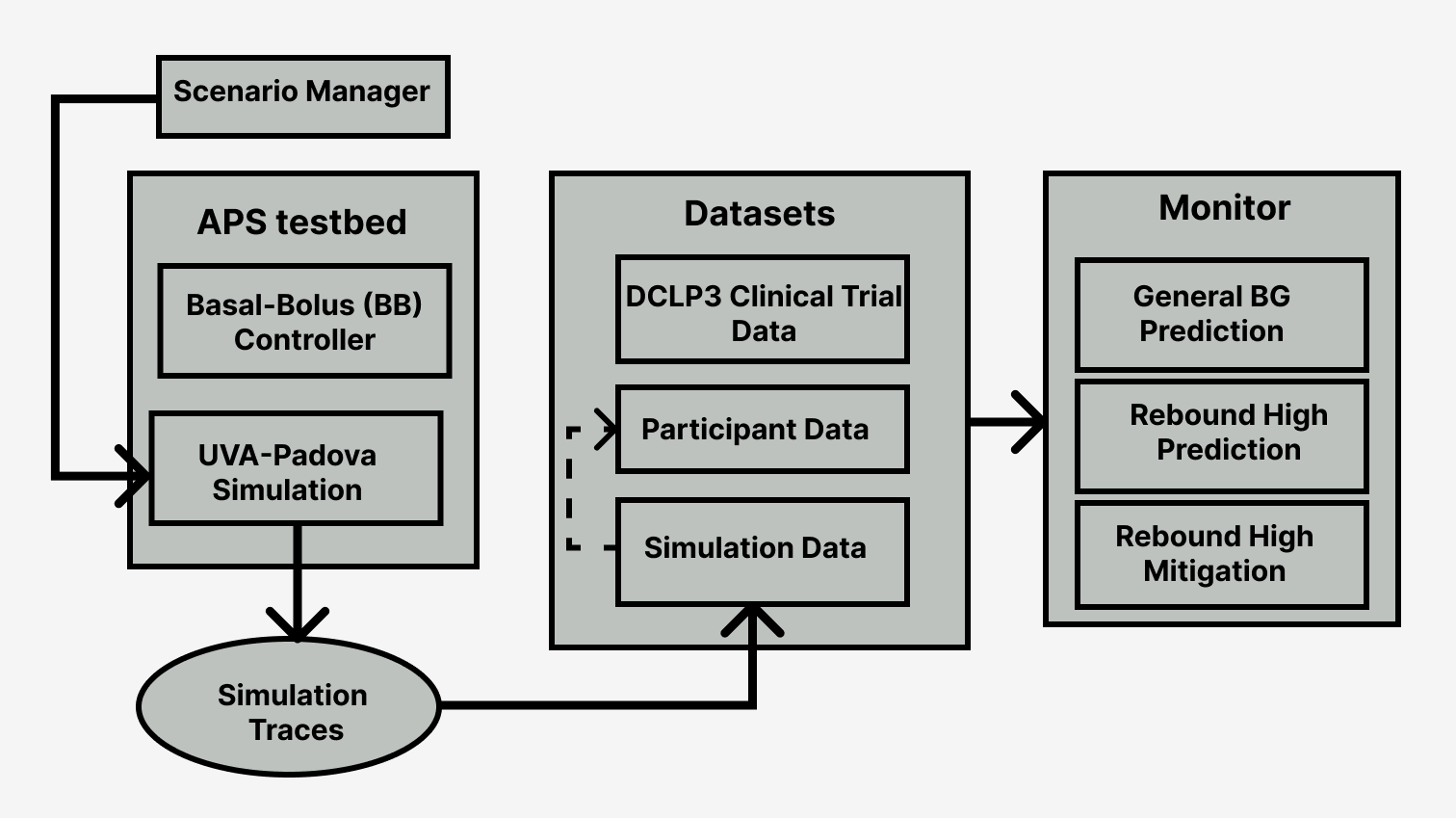} \\
        \vspace{-1em}
        \caption{Experimental Evaluation with Realistic Testbeds and Datasets}
        \label{eval}
        \vspace{-2em}
    \end{figure}
As shown in Fig. \ref{eval}, 
we use three different datasets for developing and evaluating our models, including (i) simulation data generated from a closed-loop APS testbed, (ii) a dataset collected from diabetes patients in a clinical trial, and (iii) APS data collected from one patient with type 1 diabetes. 

\textbf{Simulation Data:} The first dataset we used was generated using the UVA/Padova Type I Diabetes Simulator integrated with an open-source APS controller (Basal-Bolus) \cite{zhou2022design,zhou2021data}. 
The simulator uses an Ordinary Differential Equation-based patient model that updates using insulin received from the pump and pre-scheduled meals. The pump insulin is determined by the controller, which sets basal rates based on the virtual patient's profile and assigns boluses after a meal. CGM readings are taken from the true patient subcutaneous BG with simulated noise. This simulates a closed-loop APS, and meals can be scheduled to occur at any simulation time.

For 10 adult virtual patients, we chose 6 initial BG values each from 80 to 180 and ran 75 simulations at each initial value. In each simulation, the environment interacts with the controller for 145 timesteps (12 hours + initial state). Each timestep represented 5 minutes in actual APS. Each simulation had a meal of a random size between 30 and 100 grams of carbs after the first hour. This yielded 65,250 samples per patient or roughly 7.5 months of data.

\textbf{Clinical Trial Data:} We also used DCLP3, a publicly-available clinical trial dataset with CGM and insulin pump data for each participant \cite{DCLP3-dataset}. We chose 5 patients with at least 6 months of data.

\textbf{Participant Data:} We also used 5 days of APS data (BG and insulin) retrieved from a patient with type 1 diabetes participating in our research. Because this is not enough data to train the model alone, we used transfer learning with testbed simulation data to supplement model learning.
\subsection{Labeling and Model Training}
During evaluation, the encoder was composed of 1 LSTM layer and the decoder was composed of the attention layer, 1 LSTM layer, and 2 dense layers (both with linear activation). Each LSTM layer and the attention layer had a hidden dimension of 64, the first dense layer had a hidden dimension of 32, and the final had 1. Training lasted for 10 epochs for Simulation data and 20 epochs for DCLP3 data using the Adam optimizer with learning rate 1e-3.

For the Simulation and DCLP3 Clinical Trial datasets, we split the data for each patient into train and test sets using an 80/20 split. For participant data, we used the same 80/20 split, but first trained the model on data from 5 different virtual patients and then fine tuned the model with a low learning rate of 1e-5 to the participant's train data. The model was trained using teacher forcing and mean squared error (MSE) loss. 

Because carb inputs from meals are recorded as one large input regardless of consumption rate, we assumed a casual eating rate of 5 CHO g/min and split the carbs from the meal across multiple timesteps. For example, if a meal of 60 g of CHO was recorded at timestep $t$, we split this into 25 g at $t$, 25 g at $t+1$, and 10 g at $t+2$. This eating rate is an approximation across different patients and different food types and is meant to help the model learn by increasing the number of inputs where the CHO value is non-zero.

To determine frequency of rebound highs in each dataset, we traversed the dataset chronologically,  keeping track of the most recent BG level below 70 $mg/dL$. For each CGM measurement above 180 $mg/dL$ and within 2 hours (24 timesteps) of the most recent BG sample below 70 $mg/dL$, we labeled a rebound high and reset the <70 sample to prevent double counting.

The model monitor generates a rebound high alert whenever a hypoglycemic reading in the input is followed by a hyperglycemic prediction in the output. To test the model's ability to generate
fitting alerts, we used the test data from the regression task. If the combined sequence of input and true BG values contained a rebound high, then the combined sequence of input and predicted BG values should also contain a rebound high. Otherwise, the alert would not be triggered. Therefore, we could use the regression test data to evaluate the accuracy of rebound high alerts as well. 

\begin{table}[b]
    \vspace{-1em}
    \centering
    \caption{BG Prediction Accuracy (Lower RMSE is better)}
    \vspace{-1em}
    \resizebox{\columnwidth}{!}{%
    \begin{tabular}{c c c c c c c} \toprule
         \multicolumn{2}{c}{\textbf{Patient ID}} & \textbf{RMSE} & \textbf{CGM RMSE} & \multicolumn{2}{c}{\textbf{Patient ID}} & \textbf{RMSE} \\
        \midrule
        
        \parbox[t]{10mm}{\multirow{10}{*}{UVA Sim.}} & 1 & 11.808 & 11.011 & \parbox[t]{10mm}{\multirow{5}{*}{DCLP3}} & 3 & 27.750\\
         & 2 & 13.539 & 11.061 && 4 & 28.546\\
         & 3 & 14.166 & 11.148 && 5 & 17.988\\
         & 4 & 12.043 & 11.105 && 6 & 32.718\\
         & 5 & 14.010 & 11.114 && 7 & 38.050\\ \cline{5-7}
         & 6 & 12.342 & 11.027 & Participant & P & 31.213\\
         & 7 & 13.338 & 11.020 \\
         & 8 & 11.875 & 10.982 \\
         & 9 & 12.390 & 11.039 \\
         \bottomrule
    \end{tabular}
    }
    \label{tab:reg_results}
    \label{reg_results}
\end{table}

\subsection{Results}
The accuracy of the BG regression model on each patient is shown in Table \ref{reg_results}. In general, the root mean squared error (RMSE) of the model is higher on clinical trial patient data than simulated data due to missing records of 
carbs, time inconsistencies, and minor physiological effects absent from the simulation. We measured virtual patient simulation sensor error by calculating the RMSE between CGM measurements and true subcutaneous BG levels. Since the model learns from CGM data and cannot reduce the random noise on its own, the CGM RMSE provides a target for model performance. We do not have a similar estimate for clinical trial data because we cannot access the patients' true internal states. 

In Table \ref{tab:comp_results} we compare our approach to the same model without carb focusing (i.e., regular attention) and without attention (i.e., encoder outputs directly to the decoder LSTM) for 5 virtual and 5 DCLP3 patients. The attention layer improved the model's RMSE on test data for all but one DCLP3 patient. Incorporating carb focusing improved model performance in all but two patients (virtual 2 and DCLP3 6). This may indicate their bodies process carbs quickly, meaning attention should not be delayed to states after 30 minutes from the meal start. We also compared the model performance on participant data without transfer learning and measured an RMSE of 35.609 which was worse than 31.213 with transfer learning.

Rebound highs were much more frequent in real patient data. All but one DCLP3 patient had over 90 rebound highs in 6 months and the participant had 3 in 5 days, while simulated patients only had 3-12 in 225 days. 
For this reason, we only use the DCLP3 patients to evaluate the rebound high alert system, as shown in Table \ref{tab:alert_metrics}. One DCLP3 patient (no. 5) had no rebound highs in the test data, so it was omitted. The model performed with >98\% accuracy and >79\% precision for all DCLP3 patients, but had a low recall score, indicating that false negatives were too frequent. Despite performing well under normal conditions, it poorly predicted data traces that specifically had high carb content and volatile BG, which align with the conditions for a rebound high, but were underrepresented in the training data. 

\begin{table}[t!]
    \centering
    \caption{Comparison of Model Performance (RMSE) with and without Carb Focusing and Attention}
    \vspace{-1em}
    \resizebox{\columnwidth}{!}{%

    \begin{tabular}{c c c c c c c} \toprule
        \multicolumn{2}{c}{\textbf{Patient ID}} & \textbf{Our Approach} & \textbf{No Carb Focus} & \textbf{No Attention} \\
        \midrule
        
        \parbox[t]{10mm}{\multirow{5}{*}{UVA Sim.}} & 1 & \textbf{11.808} & 17.860 & 13.438 \\
         & 2 & 13.539 & \textbf{12.229} & 32.228 \\
         & 3 & \textbf{14.166} & 21.616 & 14.329\\
         & 4 & \textbf{12.043} & 23.589 & 70.644\\
         & 5 & \textbf{14.010} & 16.381 & 14.350\\
         \midrule
         \parbox[t]{10mm}{\multirow{5}{*}{DCLP3}} & 3 & \textbf{27.750} & 31.105 & 33.790\\
         & 4 & 28.546 & 29.310 & \textbf{27.133}\\
         & 5 & \textbf{17.988} & 18.140 & 18.119\\
         & 6 & 32.718 & \textbf{27.882} & 28.580\\
         & 7 & \textbf{38.050} & 66.761 & 60.748\\
         \bottomrule
    \end{tabular}
    }
    \label{tab:comp_results}
    \vspace{-1.5em}
\end{table}

\begin{table}[h]
    \vspace{-1em}
    \caption{Rebound High Alert System Performance Metrics}
    \vspace{-1em}
    \resizebox{\columnwidth}{!}{%
    \begin{tabular}{cccccc} \toprule
        \textbf{DCLP3} & \textbf{Rebound High} & \textbf{Accuracy} & \textbf{Precision} & \textbf{Recall} & \textbf{F1 Score} \\
        \textbf{Patient}&\textbf{Alerts to Issue}&&&& \\
        \midrule
        3 & 148 & 98.3 & 96.0 & 21.7 & 35.4 \\
        4 & 221 & 99.1 & 79.4 & 36.5 & 50.0 \\
        6 & 231 & 98.8 & 89.9 & 46.3 & 61.1 \\
        7 & 221 & 98.6 & 96.9 & 28.5 & 44.1 \\
        \bottomrule
    \end{tabular}
    }
    \vspace{-1.5em}
    \label{tab:alert_metrics}
\end{table}

\section{Conclusion and Future Work}
This paper proposes expanding the capacities of APS  controllers to predict and prevent rebound highs. Our ML BG prediction method provides scenario-specific alert messages and suggests increased basal thresholds based on carb input when a rebound high is predicted. Experimental results show that the proposed method achieves >98\% accuracy and >79\% precision for all DCLP3 patients.
Future work will adaptively choose the IB amount at run-time and include carb intake recommendations beyond standard insulin suspension in the case of rebound lows.
\section*{Acknowledgment}
This work was partially supported by Virginia CCI Grant.

\bibliographystyle{ACM-Reference-Format}
\bibliography{main}

\end{document}